\documentclass{article}

\PassOptionsToPackage{numbers}{natbib}

\usepackage[preprint]{neurips_2021}

\newif\ifaddchecklist

\usepackage[utf8]{inputenc} 
\usepackage[T1]{fontenc}    
\usepackage{hyperref}       
\usepackage{url}            
\usepackage{booktabs}       
\usepackage{amsfonts}       
\usepackage{nicefrac}       
\usepackage{microtype}      
\usepackage{xcolor}         
\usepackage{amssymb}
\usepackage{graphicx}
\usepackage{subfigure}
\usepackage{amsmath}
\usepackage{comment}
\usepackage{enumitem}

\usepackage{bbm}
\usepackage{amsthm}

\DeclareMathOperator*{\argmax}{arg\,max}
\DeclareMathOperator{\Tr}{Tr}
\usepackage{textcomp}

\title{When and how epochwise double descent happens}

\author{%
  Cory Stephenson\thanks{Work done at Intel Labs}\\
  \texttt{cory.r.stephenson@gmail.com} \\
  \And
  Tyler Lee\\
  Intel Labs\\
  \texttt{tyler.p.lee@intel.com}
}

\begin{document}

\maketitle

\begin{abstract}
Deep neural networks are known to exhibit a `double descent' behavior as the number of parameters increases. Recently, it has also been shown that an `epochwise double descent' effect exists in which the generalization error initially drops, then rises, and finally drops again with increasing training time. This presents a practical problem in that the amount of time required for training is long, and early stopping based on validation performance may result in suboptimal generalization. In this work we develop an analytically tractable model of epochwise double descent that allows us to characterise theoretically when this effect is likely to occur. This model is based on the hypothesis that the training data contains features that are slow to learn but informative. We then show experimentally that deep neural networks behave similarly to our theoretical model. Our findings indicate that epochwise double descent requires a critical amount of noise to occur, but above a second critical noise level early stopping remains effective. Using insights from theory, we give two methods by which epochwise double descent can be removed: one that removes slow to learn features from the input and reduces generalization performance, and another that instead modifies the training dynamics and matches or exceeds the generalization performance of standard training. Taken together, our results suggest a new picture of how epochwise double descent emerges from the interplay between the dynamics of training and noise in the training data.
\end{abstract}

\section{Introduction}
Training deep neural networks (DNNs) is typically a time consuming and computationally expensive process owing to the large number of parameters, non-linearity, and the resulting use of slow to converge first order optimization algorithms (SGD and its variants) \cite{ruder2016overview, du2018gradient}. Fortunately, good solutions are often obtained well before actual minimization of the objective function occurs, found either by early stopping (ES) using a validation data set, or by training until diminishing returns in generalization performance are no longer worth additional training \cite{bengio2012practical}. It is therefore of practical interest to determine when these strategies may be employed without a loss in generalization performance. 

In practice the number of training iterations is often estimated by monitoring generalization performance on validation data under the assumption that if the generalization performance no longer improves, or improves at a very slow rate then further training is unlikely to be worthwhile. However, it has been shown recently \cite{nakkiran2019deep} that this is not always the case, owing to a phenomenon called `Epochwise Double Descent' (EDD) in which the generalization performance initially improves, then begins to decrease before reversing course and improving again. This behavior mirrors the more familiar double descent effect \cite{belkin2019reconciling} seen in model complexity. If the generalization performance follows the EDD behavior, simple heuristics for determining the end of training may give sub-optimal results. For example, a common early stopping criterion is to end training if the generalization performance does not improve for a set (typically fairly small) number of training epochs. Such a heuristic will likely only capture the first descent and miss the second, even if training longer would be beneficial.

The goal of this work is to characterize the dependence of Epochwise Double Descent on the size of the model and the amount of noise in the data, and determine when this phenomenon is or is not likely to occur. To do this we use a combination of experiments and theory, and extend a line of work on double descent in linear models \cite{opper1990ability, le1991eigenvalues, krogh1992generalization, watkin1993statistical, opper1995statistical, advani2017high, mei2019generalization} to enable analysis of epochwise double descent in a theoretically tractable setting. Our main contributions can be summarized as follows:
\begin{enumerate}
    \item We introduce a solvable model that exhibits Epochwise Double Descent and find its phase diagram, showing when it occurs and when early stopping gives optimal generalization.
    \item In agreement with our theory, We show experimentally that Epochwise Double Descent occurs only if the amount of label noise exceeds a critical threshold, and simple early stopping heuristics fail only for intermediate amounts of noise.
    \item Using insights gained from our analysis, we give two modifications to the training procedure that each remove Epochwise Double Descent. One comes at a price of reduced generalization, the other matches or exceeds the generalization of standard training.
\end{enumerate}

On a practical level, our findings give insight into when simple heuristics may be used to determine when to stop training, and we provide a means of removing the epochwise double descent should it occur. From a more fundamental perspective, our findings have some tension with the generality of previously proposed `effective model complexity' (EMC) hypothesis \cite{nakkiran2019deep}. The EMC is related to the number of training samples a given training procedure can learn, and so it increases with increasing training time in the case of DNNs trained with SGD. The EMC hypothesis holds that double descent occurs as the training time is increased and the EMC exceeds the size of the training set. In contrast, we provide a training procedure that doesn't exhibit an epochwise double descent. Instead, we hypothesize an alternate explanation of the phenomenon in the presence of informative, but slow to learn features which only become relevant late in training. We provide some evidence supporting this view by showing that removal of certain features in the input can also remove the epochwise double descent phenomenon, but this comes at a cost of reducing the generalization performance.

\section{Related work}
The non-monotonic behavior of generalization error as a function of model and sample size in some types of linear models has been known since at least the late 1980s \cite{vallet1989linear} and was observed theoretically and experimentally in several works in the early 1990s \cite{opper1990ability, le1991eigenvalues, krogh1992generalization, watkin1993statistical, opper1995statistical} that studied learning using tools from statistical physics. These works connect the generalization properties of the pseudoinverse solution for perceptron-based classifiers \cite{vallet1989linear} to the eigenvalue distribution of the input covariance matrix. For i.i.d. normally distributed input data in the limit that the number of datapoints $P$ and input features $N$ are large ($P\to\infty$ and $N\to\infty$) but the ratio $\alpha=P/N$ is a finite constant, the eigenvalue distribution is given by the Marchenko–Pasteur distribution \cite{marvcenko1967distribution}.  As a result, the generalization error of the pseudoinverse solution exhibits a phase transition at $\alpha=1$ characterized by poor generalization in the vicinity of $\alpha=1$. The generalization error of this solution continues to improve for increasing $\alpha$ provided $\alpha > 1$, and exhibits a U-shaped behavior for $0<\alpha<1$ with poor generalization performance at $\alpha = 0$ and $\alpha = 1$ and better performance in between.

The peak in generalization error as the model transitions from underparameterized to overparameterized was rediscovered and phrased in the language of the bias-variance tradeoff in \cite{belkin2019reconciling}, which also named the phenomenon 'double descent.' While the connection to earlier work is unclear \cite{loog2020brief, belkin2020reply}, this work invokes the small norm bias of many common training procedures to show that the norm of the solution decreases as the amount of overparameterization increases, which is empirically associated with improvements in generalization performance \cite{belkin2018understand}. Experimentally they show that double descent occurs across many types of models, including random feature models, decision tree ensembles, and neural networks, and argue that it is a general phenomenon. However, \cite{belkin2019reconciling} notes that experiments involving neural networks trained with (stochastic) gradient descent are difficult due to the lack of theoretical solutions, long training times, and sensitivity to weight initialization.

Double descent in deep neural networks was experimentally investigated in \cite{nakkiran2019deep}. In this setting, double descent is observed as a function of the number of parameters and the size of the dataset (see the follow up work \cite{nakkiran2019more}), a finding compatible with earlier results \cite{opper1990ability, le1991eigenvalues, krogh1992generalization, watkin1993statistical, opper1995statistical}. It is also shown that early stopping removes the double descent peak, similar to the results obtained for linear networks in \cite{advani2017high, mei2019generalization}. Intriguingly, double descent as a function of the number of SGD iterations is also observed in that the test error has a peak near when the training error reaches zero, but decreases with further training. To unify these findings, this work builds from \cite{kalimeris2019sgd} and defines the 'Effective Model Complexity' (EMC) as the maximum number of samples, given a dataset and training procedure, that a model can fit with error $\leq\text{const.}$. This increases with training time as well as the number of parameters in a given model. Hypothesis 1 of \cite{nakkiran2019deep} proposes that double descent occurs generally as a function of the EMC (EMC hypothesis).

Particularly relevant for the case of deep neural networks, which typically operate deep into the overparameterized regime \cite{zhang2016understanding}, is the part of the hypothesis in \cite{nakkiran2019deep} which states (paraphrasing) that if the EMC is much larger than the size of the training dataset, any modification to the training procedure that increases the EMC results in a decrease in the test error. While supported by the experiments of \cite{belkin2019reconciling, nakkiran2019deep} in the parameter ranges considered, we note that this is different from what is known in the case of linear models, in which a minimum of the test error is achieved at a finite ratio of features to datapoints, after which increasing the number of features for a fixed amount of data results in an increasing test error.

Exploring the EMC hypothesis, \cite{heckel2020early} finds that in the case of linear regression, epochwise double descent occurs as a consequence of the gradient descent training dynamics when differently scaled features are learned at different times. In this setting, the test risk decomposes into a sum of bias-variance tradeoffs which occur at different times, resulting in an epochwise double descent. This effect can be removed by taking an appropriate learning rate for each feature, which lines up the different bias-variance tradeoffs in time, an effect which sometimes can improve generalization. Additionally, \cite{heckel2020early} shows analytically that similar effects occur in a two layer network, and experimentally demonstrates similar behavior in a 5-layer convolutional network. 

Our work aims to bring our understanding of epochwise double descent closer to that of double descent with model complexity. Specifically, we characterize how the amount of data and noise interacts to produce an epochwise double descent given generic assumptions on the training data.

\section{Setup for analysis of training dynamics}
\label{sec:setup}
We consider training a model on $N$ samples $\{(x_1, y_1), (x_2, y_2), ..., (x_N, y_N)\}_{x,y\sim\mathcal{D}_{train}}$ where $x_i\in\mathbb{R}^D$, and $y_i\in\mathbb{R}^C$. That is, a dataset of $N$ samples from a distribution $\mathcal{D}$ with $D$ input features and $C$ classes. For convenience, we adopt the bold notation for matrices, and group the input features into the data matrix $\mathbf{X}\in\mathbb{R}^{D\times N}$, and similarly, the matrix $\mathbf{\Phi}\in\mathbb{R}^{F\times N}$ given by the result of an $F$ dimensional embedding of $x$ by $\left[\mathbf{\Phi}\right]_{ij}=\phi_i(x_j)$ where each $\phi_i(\cdot)$ is a map $\phi_i:\mathbb{R}^D\to\mathbb{R}$. Let $\mathbf{y}\in\mathbb{R}^{C\times N}$, and $\mathbf{\hat{y}}\in\mathbb{R}^{C\times N}$ be the training target matrix and model output matrix respectively. In the case of probabilistic (hard) labels we take $\left[\mathbf{y}\right]_i =\left[\mathbf{P_L}\right]_i\in\{0, 1, ..., C\}$.

We first consider training with mean squared error (MSE) cost $\mathcal{L}_{MSE}$ defined as
\begin{equation}
    \mathcal{L}_{MSE} = \frac{1}{2N}\Tr{\left[\left(\mathbf{w}\mathbf{\Phi} - \mathbf{y}\right)^\intercal\left(\mathbf{w}\mathbf{\Phi} - \mathbf{y}\right)\right]}
\end{equation}
For some weights $\mathbf{w}\in\mathbb{R}^{C\times F}$. Gradient descent on this cost with a learning rate of $\gamma$ gives the dynamics
\begin{equation}
    \label{eqn:mse-dynamics}
    \mathbf{w}_{MSE}^{(t+1)} = \mathbf{w}_{MSE}^{(t)} - 
    \frac{\gamma}{N}\mathbf{w}_{MSE}^{(t)}\mathbf{\Phi}\mathbf{\Phi}^\intercal + \frac{\gamma}{N}\mathbf{y}\mathbf{\Phi}^\intercal
\end{equation}
The MSE dynamics permit an analytic solution (see appendix for the derivation)
\begin{equation}
    \label{eqn:mse-dynamics-solution}
    \mathbf{w}_{MSE}^{(t)} = \mathbf{w}_{MSE}^{(\infty)} + \left(\mathbf{w}_{MSE}^{(0)} - \mathbf{w}_{MSE}^{(\infty)}\right)\mathbf{U}\left[\mathbb{I}_F - \gamma\mathbf{\Lambda}\right]^t\mathbf{U}^\intercal
\end{equation}
Where we have used the decomposition $\mathbf{\Phi}\mathbf{\Phi}^\intercal/N=\mathbf{U}\mathbf{\Lambda}\mathbf{U}^\intercal$. The parameters after infinite training time are given by
\begin{equation}
    \mathbf{w}_{MSE}^{(\infty)} = \mathbf{y}\mathbf{\Phi}^\intercal\left(\mathbf{\Phi}\mathbf{\Phi}^\intercal\right)^+ + \mathbf{w}_{MSE}^{(0)}\left(\mathbb{I}_{F} - \mathbf{U}\mathbf{\Lambda}^+\mathbf{\Lambda}\mathbf{U}^\intercal\right)
\end{equation}
Where the superscript $+$ denotes the Moore-Penrose pseudoinverse and $\mathbb{I}_F$ is the $F\times F$ identity.

In the appendix we show that in the high temperature limit the dynamics of training with softmax cross entropy permits a similar solution for the trajectory of the paramters $\mathbf{w}_{XENT}$
\begin{equation}
    \mathbf{M}\mathbf{w}_{XENT}^{(t)} = \mathbf{M}\mathbf{w}_{XENT}^{(\infty)} + \mathbf{M}\left(\mathbf{w}_{XENT}^{(0)} - \mathbf{w}_{XENT}^{(\infty)}\right)\mathbf{U}\left[\mathbb{I}_F - \gamma\mathbf{\Lambda}\right]^t\mathbf{U}^\intercal
\end{equation}
After defining $\mathbf{M} \equiv \mathbb{I}_C - \frac{1}{C}\mathbbm{1}_{CC}$, and introducing $\mathbbm{1}_{CN}$ as a $C\times N$ matrix of ones. The parameters after infinite training time are given similarly:
\begin{equation}
        \mathbf{M}\mathbf{w}_{XENT}^{(\infty)} = \mathbf{M}\left(C\mathbf{P_L}-\mathbbm{1}_{CN}\right)\mathbf{\Phi}^\intercal\left(\mathbf{\Phi}\mathbf{\Phi}^\intercal\right)^+ + \mathbf{M}\mathbf{w}_{XENT}^{(0)}\left(\mathbb{I}_{F} - \mathbf{U}\mathbf{\Lambda}^+\mathbf{\Lambda}\mathbf{U}^\intercal\right)   \label{eqn:softmax-final-weights}
\end{equation}
This is nearly identical to the solution for MSE given in Eq. \ref{eqn:mse-dynamics-solution} if the label matrix in the MSE case is identified as $\mathbf{y}=\left(C\mathbf{P_L}-\mathbbm{1}_{CN}\right)$. The matrix $\mathbf{M}$ acts to subtract the mean across classes of whatever it acts on, so the cross entropy and MSE solutions are identical other than differences in this mean. This relationship allows us to gain insight into the generalization behavior of training with softmax cross entropy by analyzing the easier MSE case.

\section{A simple model of epochwise double descent}
\label{sec:toy-model}

\begin{figure}[t!]
\centering
\includegraphics[width=0.9\columnwidth]{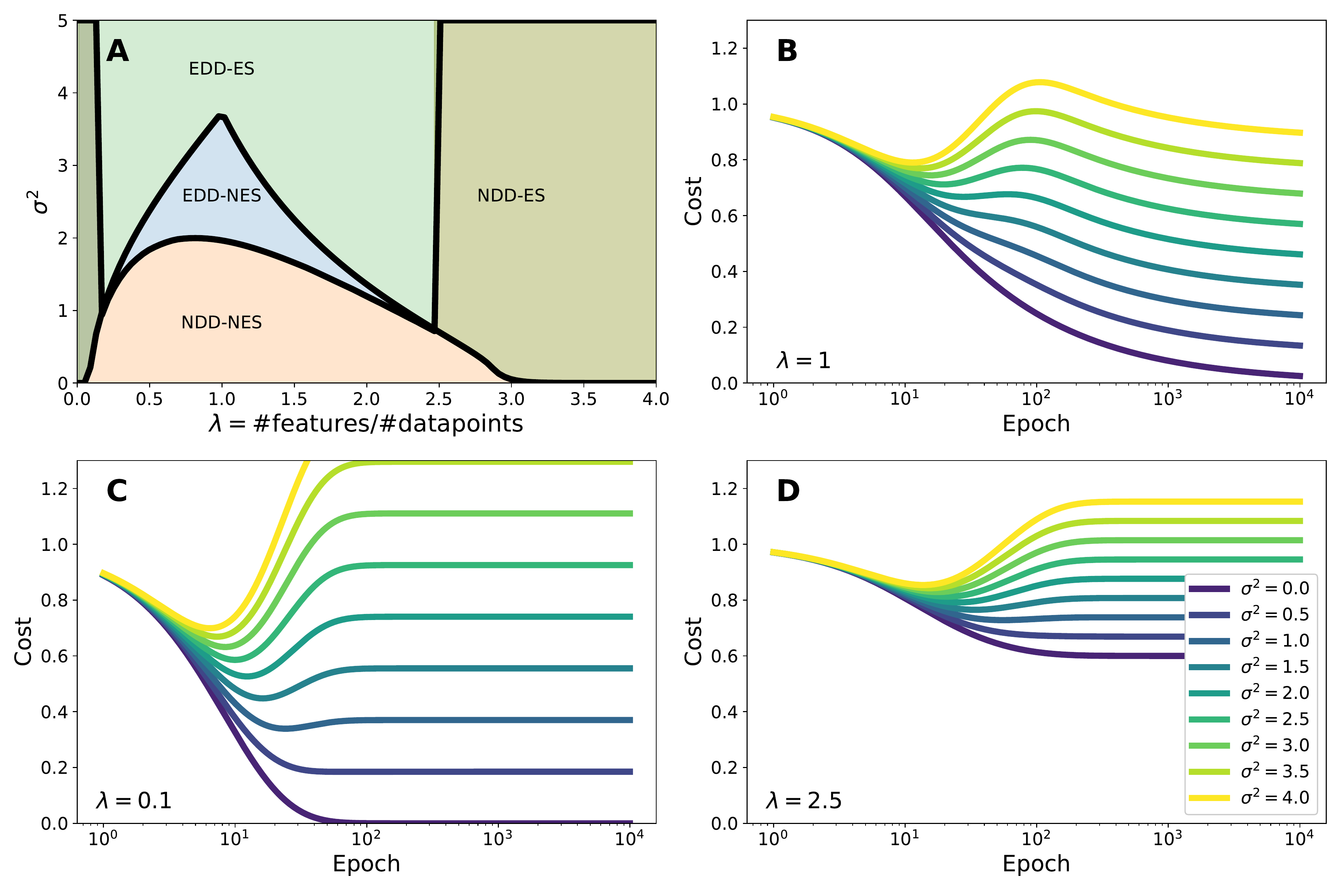}
\caption{\textbf{A: }Phase diagram in the $\lambda, \sigma^2$ plane. The blue region indicates where epochwise double descent occurs and the second descent gives the best generalization performance. The light green region indicates where epochwise double descent occurs but the first descent gives the best generalization performance. \textbf{B: } Generalization error vs. training epoch at critical parameterization. Higher values of noise produce epochwise double descent. \textbf{C: }Generalization error vs. training epoch in the underparameterized regime. Early stopping is necessary for optimal performance at larger amounts of noise. \textbf{D: } Generalization error vs. training epoch in the overparameterized regime. Early is necessary for optimal performance at larger amounts of noise, but is unnecessary for small amounts of noise}
\label{fig:phase-diagram}
\end{figure}

\begin{figure}[t!]
\centering
\includegraphics[width=0.99\columnwidth]{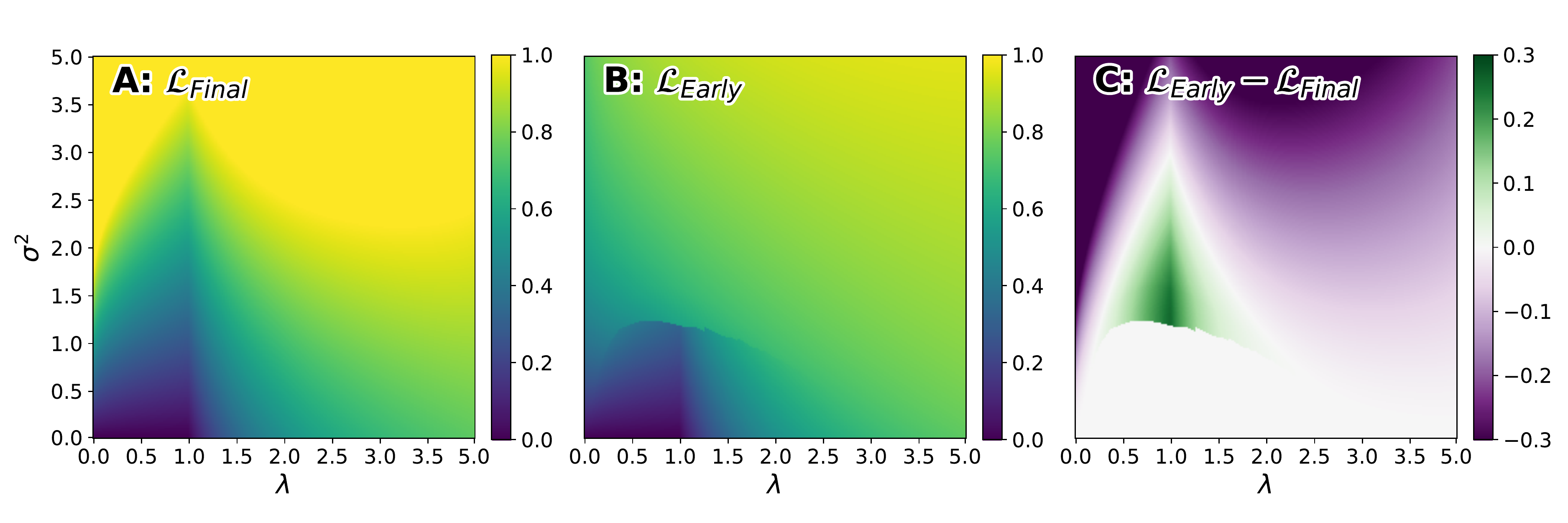}
\caption{\textbf{A: } Generalization error at convergence ($t\to\infty$)in the $\lambda, \sigma^2$ plane. \textbf{B: } Generalization error at the early stopping time in the $\lambda, \sigma^2$ plane. \textbf{C: } Difference in generalization error between the early stopping time and convergence. Green indicates early stopping gives suboptimal generalization performance due to epochwise double descent.}
\label{fig:phase-diagram-heatmap}
\end{figure}

In building a toy model of epochwise double descent, we note that the gradient descent training dynamics shown in Sec. \ref{sec:setup} is such that large eigenvalues of the covariance matrix are learned first, and this has implications for observing epochwise double descent in linear models. In particular, the small eigenvalues will dominate the learning at long training times, and if test error is to decrease in this regime, the small eigenvalues must be relatively noise free. However, if the test error is to increase at intermediate training times, the larger eigenvalues must be noisy. In many analyses, such as \cite{advani2017high}, the noise is typically assumed to be uniform across eigenvalues, and so epochwise double descent does not occur in these works. Here we construct a model in which the noise only has an effect on the large eigenvalues of the data covariance matrix, which is sufficient to produce an epochwise double descent effect near $\alpha=1$.

We assume a training dataset $\mathbf{X}\in\mathbb{R}^{N\times D}$ consisting of $N$ datapoints each with $D$ normally distributed features such that $\left[\mathbf{X}\right]_{i, j}\sim\mathcal{N}(0, 1)$. These data points are assigned labels $\mathbf{y}\in\mathbb{R}^{C\times N}$ via $\mathbf{y}=\mathbf{w}_T\mathbf{X} +\epsilon$ labeled using a teacher matrix $\mathbf{w}_T\in\mathbb{R}^{C\times D}$ chosen such that $\left[\mathbf{w}_T\right]_{i,j}\sim\mathcal{N}(0, 1/\sqrt{D})$. Here, $\epsilon$ is a label noise term, but we do not assume it is i.i.d. normal. Instead, we assume $\epsilon$ couples to eigenvectors with large eigenvalues only. Using the SVD $\mathbf{X}/\sqrt{N}=\mathbf{U}\Lambda^{1/2}\mathbf{V}^\intercal$, we define $\mathbf{z} = \epsilon\mathbf{V}$ such that $\mathbf{z}\in\mathbb{R}^{C\times N}$, or equivalently, $\epsilon=\mathbf{z}\mathbf{V}^\intercal$, and choose $\mathbf{z}$ according to the following:
\begin{equation}
    \label{eqn:noise-definition}
    \left[\mathbf{z}\right]_{ij} \sim    \begin{cases}
      \mathcal{N}(0, \sigma) & \left[\mathbf{\Lambda}\right]_{jj} \geq 1 \\
      0 & \left[\mathbf{\Lambda}\right]_{jj} \leq 1\\
    \end{cases}
\end{equation}
Note that we still have $\left<\epsilon\right>=0$ and $\left<\epsilon\mathbf{X}^\intercal\right>=0$, however the noise is now correlated across examples. For a further discussion on this type of noise, and how it relates to randomly labeled examples in a typical training setup, see the appendix.

The eigenvalues of the data covariance matrix $\mathbf{XX^\intercal}/N$ follow the Marchenko-Pasteur (MP) distribution \cite{marvcenko1967distribution} in the limit $N, D \to\infty$ with $D/N=\lambda$.
\begin{equation}
    p_{MP}(x\vert\lambda) = \frac{1}{2\pi}\frac{\sqrt{(\lambda_+ - x)(x - \lambda_-)}}{\lambda x}\mathbf{1}_{x\in\left[\lambda_-, \lambda_+\right]}
\end{equation}
with $\lambda_\pm=(1\pm\sqrt{\lambda})$. Note that there is a point mass of $p_{MP}(0) = 1 - 1/\lambda$ at $x=0$ if $\lambda>1$. Additionally, the discontinuity at $\left[\Lambda\right]_{jj}=1$ in Eq. \ref{eqn:noise-definition} corresponds to the mean of the MP distribution. With this distribution and the solutions in Sec. \ref{sec:setup} we can calculate the expected generalization error (as measured by the MSE cost function) after $t$ iterations of training as well as it's dependence on the amount of noise, measured by $\sigma$ and the amount of over/under parameterization as measured by $\lambda$. The underparameterized regime has $\lambda<1$ and the overparameterized regime has $\lambda> 1$.
\begin{equation}
    \label{eqn:generalization-error}
    \begin{aligned}
    \left<\mathcal{L}_{test}(t, \lambda, \sigma)\right>_{\epsilon, \mathbf{X}, \mathbf{w}_T} =& \int_{1}^{\infty}\left[\left(1 + \frac{\sigma}{x}\right)\left(1-\gamma x\right)^{2t} + \frac{\sigma}{x}\left(1 - 2(1-\gamma x)^{t}\right)\right]p_{MP}(x\vert\lambda)dx\\
    &+ \int_{0}^{1}\left[(1-\gamma x)^{2t} \right]p_{MP}(x\vert\lambda)dx
    \end{aligned}
\end{equation}
Where the average has been taken over the teacher matrix, the training data, and the noise.

The dynamics given by Eq. \ref{eqn:generalization-error} can be qualitatively different depending on the amount of training data and the amount of noise. We define four different classes of behavior depending on whether or not an epochwise double descent occurs and whether or not the minimum of the generalization cost occurs before the end of training, necessitating early stopping. we define:
\begin{enumerate}
    \item\textbf{NDD-NES: No epochwise double desecnt, no early stopping}: Here the generalization cost decreases monotonically with increasing training time. We observe this behavior for small amounts of noise especially near critical parameterization $\lambda\approx1$.
    \item\textbf{NDD-ES: No epochwise double descent, early stopping}: Here the generalization cost first decreases, then increases and plateaus above it's minimum value, meaning early stopping is necessary for optimal generalization performance.
    \item\textbf{EDD-NES: Epochwise double descent occurs, no early stopping}: Here, epochwise double descent occurs in that the generalization cost first decreases, then increases, and then decreases again. The second decrease is large enough to overcome the increase, and so early stopping gives sub-optimal generalization.
    \item\textbf{EDD-ES: Epochwise double descent occurs, early stopping is necessary}: Here, epochwise double descent occurs in that the generalization cost first decreases, then increases, and then decreases again. The second decrease is not large enough to overcome the increase, and so early stopping is necessary.
\end{enumerate}

Each of these types of behaviors occurs in this model, and the situation is summarized in Fig. \ref{fig:phase-diagram}. The generalization error obtained by early stopping and training to convergence is shown in Fig. \ref{fig:phase-diagram-heatmap}. We observe a discontinuous drop in generalization performance when using early stopping at the onset of epochwise double descent in Fig. \ref{fig:phase-diagram-heatmap}B. We identify four properties P1-P4 with practical implications:
\begin{enumerate}[label=P{{\arabic*}}]
    \item For clean datasets (small $\sigma$), early stopping is not crucial for achieving good generalization performance.
    \item Epochwise double descent only occurs near critical parameterization.
    \item Epochwise double descent requires noise above a certain threshold to occur.
    \item If the amount of noise is too large (large $\sigma$), early stopping gives the best generalization performance, and epochwise double descent can be ignored in practice.
    \label{properties-list}
\end{enumerate}
The first of these is known to hold true in deep neural networks as well, as common practice is to train models for as long as practically feasible regardless of the possibility of overfitting. The second of these has been partially demonstrated in \cite{nakkiran2019deep} which shows that networks below a critical size do not exhibit epochwise double descent.

\section{EDD is only relevant within a specific noise range}
\label{sec:width-sweep}

\begin{figure}[t!]
\centering
\begin{minipage}{1\textwidth}
\centering
\includegraphics[width=1\columnwidth]{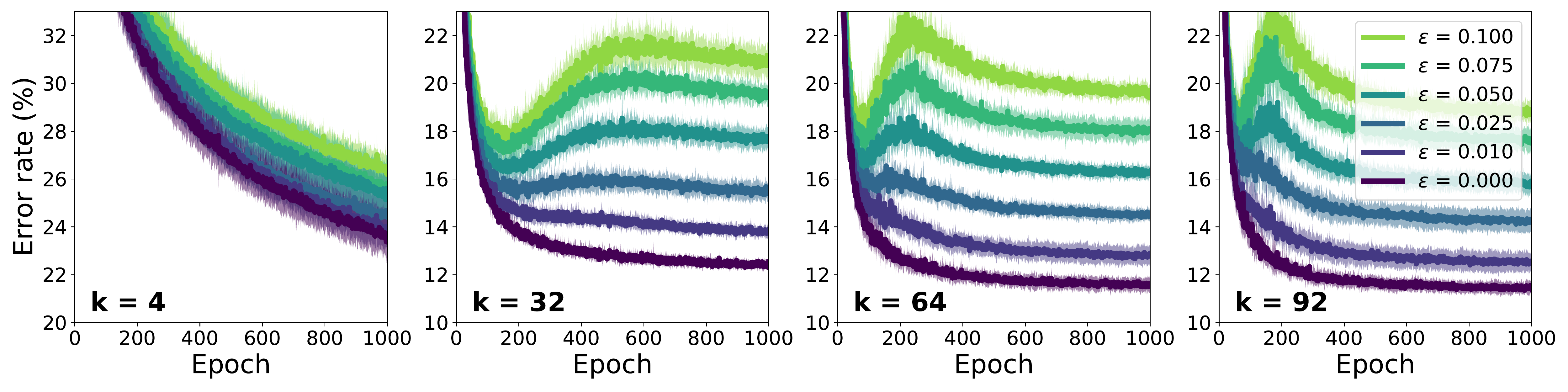}
\end{minipage}
\caption{Generalization behavior of ResNet18 trained on CIFAR10 at various widths and noise levels. Larger values of $k$ indicate a higher total parameter count. Epochwise double descent occurs at larger values of $k$ at noise levels above $\epsilon\approx0.025$. At larger values of the noise $\epsilon$ the second descent does not achieve a lower generalization error than the first descent. Note the different scale on $k=4$. All plots show mean (bold) $\pm$ std (shaded) across 10 random seeds.}
\label{fig:width-sweep}
\end{figure}

To evaluate properties 1-4 extracted from the linear model of Sec. \ref{sec:toy-model} we replicated the experimental setup of \cite{nakkiran2019deep} and trained instances of ResNet18 \cite{he2016deep} with SGD and varying widths $k$ on the CIFAR10 dataset \cite{krizhevsky2009learning}. This work was implemented using PyTorch \cite{NEURIPS2019_9015} and the TorchVision ResNet18 implementation. We did not replicate the experiments on CIFAR100 \cite{krizhevsky2009learning} as clean, unambiguous labels are required to investigate the generalization behavior with small amounts of label noise. CIFAR100 is known to have a much larger amount of label noise/ambiguity than CIFAR10 \cite{northcutt2021pervasive}. We investigated noise fractions ranging from $\epsilon=0.0$ to $\epsilon=0.1$ random label noise, and averaged the test error across 10 random seeds\footnote{This random seed determines the selection and relabling of the randomly labeled examples as well as the network initialization} for each combination of width and amount of label noise.

Results can be seen in Fig. \ref{fig:width-sweep}. At $\epsilon=0$, test error monotonically decreases for all network widths, and so there is no well defined early stopping epoch (property 1). For smaller networks widths ($k=4$ and $k=32$) we observe little to no epochwise double descent at any noise levels, a finding consistent with \cite{nakkiran2019deep} (property 2). At larger widths, such as the standard ResNet18 ($k=64$) and larger ($k=92$) we see a strong epochwise double descent effect above $\epsilon=0.025$ ($2.5\%$ noisy labels). At $\epsilon=0.025$ a small double descent peak begins to emerge, though it is similar in magnitude to the jitter in the test error due to SGD. Below $\epsilon=0.025$, we do not observe epochwise double descent at any width. This is consistent with the phase diagram in Fig. \ref{fig:phase-diagram}A in the NDD-NES regime which suggests there is a critical noise level below which epochwise double descent does not occur (property 3).

At larger widths ($k=64, 92$) and larger noise values ($\epsilon=0.1$), we observe a strong epochwise double descent effect. However, the initial descent achieves a lower test error than the second descent, and so a simple early stopping heuristic outperforms training to convergence. This is also consistent with the EDD-ES regime in the phase diagram in Fig. \ref{fig:phase-diagram}A in which epochwise double descent occcurs but can be safely ignored via early stopping (property 4). For large widths and intermediate noise levels ($\epsilon=0.025-0.05$ for $k=64$ and $\epsilon=0.025-0.075$ for $k=92$) epochwise double descent occurs, and the lowest test error is achieved at the end of training. This is the regime in which early stopping gives sub-optimal performance, similar the EDD-NES regime in Fig. \ref{fig:phase-diagram}A. 

Our linear model also predicts that networks which are \textit{above} a certain size will also not exhibit epochwise double descent. While this may happen for very large networks, we were not able to observe it experimentally in ResNet18. Experiments at large values of $k$ are quite computationally expensive\footnote{The number of parameters grows like $k^2$}, so it is possible our networks are simply not large enough to show this effect. Data augmentation and parameter redundancy both act to reduce the number of effective parameters, so in the case of ResNet18 it is unclear how large $k$ should be to reach the correct regime. We note that this occurs in the same regime as the decrease in generalization performance seen with increasing overparameterization regime shown in linear models (\cite{advani2017high} and others) which has similarly not been observed in deep neural networks.

\section{Removing epochwise double descent by removing features}
\label{sec:pca-experiment}
\begin{figure}[t!]
\centering
\label{fig:pca-training}
\includegraphics[width=0.99\columnwidth]{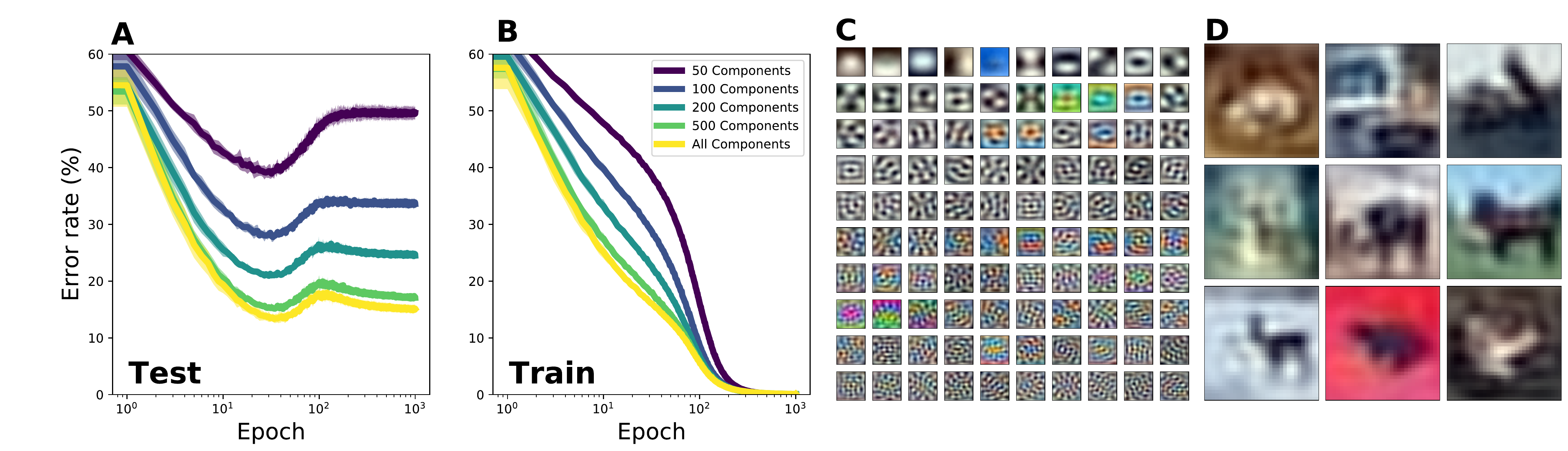}
\caption{\textbf{A: } Generalization error of ResNet18 trained on CIFAR10 with 20\% label noise ($\epsilon=0.2$) as an increasing number of the smaller principal components are removed. When $\leq100$ components remain, epochwise double descent disappears. \textbf{B: } Training error as an increasing number of the smaller principal components are removed. \textbf{C: } Visualization of the first 100 principal components showing that the larger principal components correspond to larger scale features. \textbf{D: } Examples of the training data when only the 100 largest principal components are kept. All lines show mean (bold) $\pm$ std (shaded) across 10 random seeds.}
\end{figure}

A central assumption of the linear model in Sec. \ref{sec:toy-model} is the existence of slow to learn but informative features. In the linear setting, these features consist of eigenvectors with small eigenvalues. In the case of DNNs, it is less clear what features in the input are slow to learn owing to the nonlinearity of the training dynamics and differing architecture choices. However, very wide DNNs are known to behave approximately linearly \cite{lee2019wide} in a transformed feature space determined by the network architecture and initial weights. While this transformation can change the ordering in which features are learned and introduce new nonlinearly constructed features, there may still be some correspondence between the slow to learn features in the linear case and the slow to learn features in a DNN. With this in mind, we carried out an experiment in which we discarded the smaller eigenvalue features of a dataset that shows epochwise double descent and trained a DNN on the remaining components.

Fig. \ref{fig:pca-training}A shows the result of training\footnote{Note: hyperparameters here are different from \cite{nakkiran2019deep} and were obtained with a hyperparameter sweep. See appendix for details} ResNet18 on CIFAR10 with $\epsilon=0.1$ and a varying number of the largest principal components, leaving the test set unmodified. All results are averaged across 10 random seeds. When all components are kept (yellow line) we see epochwise double descent. However, as the number of components is reduced the epochwise double descent phenomenon becomes less pronounced, and vanishes when only the $100$ largest components (which account for 90\% of the variance) are kept. This is consistent with the features responsible for the epochwise double descent having been removed. As the removed features are informative, this also corresponds to a decrease in generalization performance. The top 100 PCA components are shown in Fig. \ref{fig:pca-training}C and the modified data with 100 components in Fig. \ref{fig:pca-training}D. As might be expected, the larger principal components tend to represent larger scale features. Fig. \ref{fig:pca-training}B demonstrates that ResNet18 is still able to fit this modified training data. 

\section{Removing epochwise double descent by training the last layer}
\label{sec:retrain-experiment}
\begin{figure}[t!]
\centering
\label{fig:retrain-last}
\includegraphics[width=1\columnwidth]{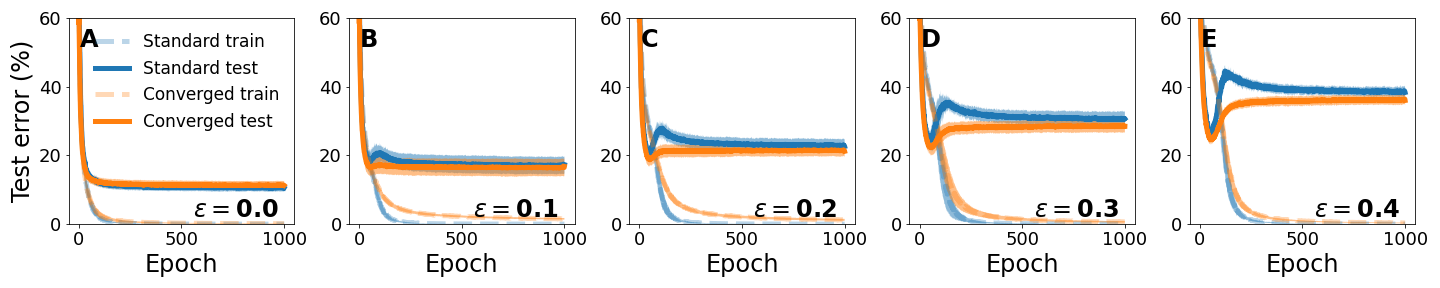}
\caption{Comparasion of training with SGD (standard training, Blue) and substituting the analytic solution for the final layer weights after training with SGD (converged, Orange). Using the analytic solution for the final layer weights removes epochwise double descent and gives equal or better generalization error. All plots show mean (bold) $\pm$ std (shaded) across 10 random seeds.}
\end{figure}

While the experiment in Sec. \ref{sec:pca-experiment} demonstrates that epochwise double descent can be eliminated by removing certain useful features in the training data, this also causes a significant drop in generalization performance. However, in the linear model of Sec. \ref{sec:toy-model} epochwise double descent appears as a property of the training dynamics as well as the training data which hints at an alternate method of removing the effect. If it were possible to skip to the end of training, the order in which features are learned becomes irrelevant and there is no possibility for an epochwise double descent to exist. In general, this is not possible for a DNN as a whole as analytic solutions don't exist, with the exception of the final layer.

As previous works have shown, representations in early layers of DNNs typically stabilize quickly \cite{raghu2017svcca, morcos2018insights}, while the final layers are responsible for memorizing noise \cite{stephenson2021on} and continue to learn during training. This leads us to suspect that training just the final layer to convergence might be sufficient to remove the epochwise double descent as this layer plays a dominant role in the later epochs of training. This can be done in the case of a network trained with softmax cross entropy using the smooth-label approximation for the weights given in Eq. \ref{eqn:softmax-final-weights}. Our procedure is as follows: Train a DNN with SGD using the softmax cross entropy objective for $t$ epochs (standard training procedure). Then, replace the weights of the final classification layer with the 'converged' weights given by Eq. \ref{eqn:softmax-final-weights} using the penultimate layer activations as features. The result is a network which has the final layer trained to convergence, and the earlier layers trained for $t$ epochs with a standard training procedure.

Fig. \ref{fig:retrain-last} shows the test error achieved by ResNet18 ($k=64$) trained\footnote{Hyperparameters are those from Sec. \ref{sec:pca-experiment}} on CIFAR10 for values of $\epsilon$ between 0.0 and 0.4. The blue line represents the error achieved with standard training, and the orange 'Converged' line represents the error achieved after substituting in the converged weights of Eq. \ref{eqn:softmax-final-weights}. We see that in all cases, using the 'converged' weights eliminates epochwise double descent. When $\epsilon > 0$, 'converged' weights give also give a lower test error than the standard training, as the second descent in test error results in a very slow convergence to this final value. In the noise free case of $\epsilon=0$, we find that standard training and using 'converged' weights for the final layer give equivalent performance, with differences smaller than variance arising from different random seeds.

\section{Discussion and Conclusion}
In constructing the linear model of epochwise double descent, we assumed the existence of small scale (small eigenvalue) features that are largely unaffected by the presence of noise, and learn slowly compared to to larger scale (large eigenvalue) features which are noisy. Near critical parameterization, this gives rise to an epochwise double descent for noise levels above a threshold. This is similar to other linear models (recently \cite{advani2017high, heckel2020early}) that achieve the well-studed double descent in model complexity by assuming the existence of small scale (small eigenvalue) which couple to uniform noise. While not a focus of this work, we note that both effects can coexist if the small scale features are \textit{weakly} affected by noise in comparison to the large scale features. We believe that the study of other noise models is likely to yield other behaviors which may be interesting, and the problem of determining which noise model is most appropriate for the case of training deep neural nets with label noise is a promising direction for future work. 

We also demonstrated that epochwise double descent can be removed by the deletion of certain features in the input data (Sec. \ref{sec:pca-experiment}). While in this work we simply discarded the smallest principal components, this is likely suboptimal. Other approaches which are more selective about which input features to remove might have the potential to remove epochwise double descent at a smaller cost to generalization performance. Determining which features are responsible for epochwise double descent may also shed light on what features DNNs are sensitive to at different stages of training, and so this is a potential avenue for further work on DNN interpretability in connection to the various double descent phenomena. 

Finally, we note an apparent tension between our experimental results and the EMC hypothesis of \cite{nakkiran2019deep}. In Sec. \ref{sec:retrain-experiment} we give an example of a training procedure that does not exhibit double descent as the number of samples fit with error $\leq\text{const.}$ exceeds the size of the training set. Strictly speaking, this acts as a counterexample to the EMC hypothesis. Furthermore, we also note that our results in Sec. \ref{sec:width-sweep} suggest that a critical amount of noise is necessary to produce epochwise double descent, while the EMC hypothesis suggests that epochwise double descent should occur for all noise levels. We provide an alternate view: epochwise double descent occurs as a result of an interplay between the features in the data and the noise in the labels (as in out linear model) rather than as a result of the EMC. We believe additional work clarifying which view of the epochwise double descent phenomena is correct would improve the understanding of the generalization dynamics of DNNs.

In summary, we have developed a linear model which exhibits an epochwise double descent effect. We have calculated a phase diagram that shows when epochwise double descent does/does not occur, and when early stopping does/does not lead to optimal generalization performance. In spite of the simplicity of our linear model, all four combinations of these two effects occur depending on the amount of overparameterization and the amount of label noise. Experiments on deep neural networks show that these highly nonlinear models behave qualitatively similarly to our linear model, in that epochwise double descent requires both overparameterization and noise above a critical threshold to occur. Experimentally and in our linear model we find that above a second critical noise level epochwise double descent occurs but is not relevant as early stopping gives superior generalization performance. Using insights from the linear model we give two methods that experimentally eliminate the epochwise double descent effect, one which harms generalization and another which improves or matches the generalization performance of standard training. These findings follow from our hypothesis that the epochwise double descent effect arises from slow to learn but informative features. We hope future works will shed further light on the nature of these features and their relation to the influence of noisy labels on the generalization dynamics of deep neural nets.

\paragraph{Broader Impacts}
Deep learning has become an incredibly compute-intensive field, where long-running model training runs are the norm. Epochwise double descent poses a challenge in this context, since, as described, early stopping may yield worse generalization performance. Beyond this, our work is application-agnostic, so we cannot foresee any clear negative impacts.

\clearpage
\small
\bibliography{references}
\bibliographystyle{unsrt}
\clearpage

\section{Appendix}
\subsection{Derivation of MSE dynamics}
The mean-squared error ($\mathcal{L}_{MSE}$) is given by 
\begin{equation}
    \mathcal{L}_{MSE} = \frac{1}{2N}\sum_{j=1}^N\sum_{c=1}^C\left[\mathbf{w}\mathbf{\Phi} - \mathbf{y}\right]_{cj}^2
\end{equation}
Computing the gradient with respect to one of the parameters $\mathbf{w}_{cj}$ gives
\begin{equation}
    \frac{\partial\mathcal{L}_{MSE}}{\partial\mathbf{w}_{cj}} = \frac{1}{N}\left[\mathbf{w}\mathbf{\Phi}\mathbf{\Phi}^\intercal - \mathbf{y}\mathbf{\Phi}^\intercal\right]_{cj}
\end{equation}
Carrying out gradient descent with a learning rate $\gamma$ gives the recursion relation for the parameters (Eq. \ref{eqn:mse-dynamics} from the main text)
\begin{equation}
    \mathbf{w}_{MSE}^{(t+1)} = \mathbf{w}_{MSE}^{(t)} - 
    \frac{\gamma}{N}\mathbf{w}_{MSE}^{(t)}\mathbf{\Phi}\mathbf{\Phi}^\intercal + \frac{\gamma}{N}\mathbf{y}\mathbf{\Phi}^\intercal
\end{equation}
Where $\mathbf{w}_{MSE}^{(t)}$ denotes the value of parameters obtained after training for time $t$ with MSE. This relation has a fixed point at
\begin{equation}
 \mathbf{w}_{MSE}^{*} = \mathbf{y}\mathbf{\Phi}^\intercal\left(\mathbf{\Phi}\mathbf{\Phi}^\intercal\right)^+
\end{equation}
where $(\cdot)^+$ denotes the Moore-Penrose pseudoinverse. Changing variables to $\mathbf{z}^{(t)} = \mathbf{w}_{MSE}^{(t)} - \mathbf{w}_{MSE}^{*}$ Gives
\begin{equation}
    \mathbf{z}^{(t+1)} = \mathbf{z}^{(t)} -\frac{\gamma}{N}\mathbf{z}^{(t)}\mathbf{\Phi}\mathbf{\Phi}^\intercal
\end{equation}
Changing variables to the eigenbasis via $\mathbf{\Phi}\mathbf{\Phi}^\intercal/N = \mathbf{U}\mathbf{\Lambda}\mathbf{U}^\intercal$ and $\tilde{\mathbf{z}}^{(t)} = \mathbf{z}^{(t)}\mathbf{U}$ gives a set of decoupled linear recursion relations
\begin{equation}
    \tilde{\mathbf{z}}^{(t+1)} = \tilde{\mathbf{z}}^{(t)} -\gamma\tilde{\mathbf{z}}^{(t)}\mathbf{\Lambda}
\end{equation}
Which can easily be solved to give
\begin{equation}
    \tilde{\mathbf{z}}^{(t)} = \tilde{\mathbf{z}}^{(0)}\left(\mathbb{I}_F - \gamma\mathbf{\Lambda}\right)^t
\end{equation}
Changing variables back to the original basis with $\tilde{\mathbf{z}}^{(t)} = \mathbf{z}^{(t)}\mathbf{U}$ and $\mathbf{z}^{(t)} = \mathbf{w}_{MSE}^{(t)} - \mathbf{w}_{MSE}^{*}$ gives the solution for the parameters
\begin{equation}
    \mathbf{w}_{MSE}^{(t)} = \mathbf{w}_{MSE}^{*} + \left(\mathbf{w}_{MSE}^{(0)} - \mathbf{w}_{MSE}^{*}\right)\mathbf{U}\left(\mathbb{I}_F - \gamma\mathbf{\Lambda}\right)^t\mathbf{U}^\intercal
\end{equation}

In the case where some eigenvalues of $\mathbf{\Phi}$ are zero, there is a frozen subspace in which the parameters don't change in time and so the infinite time solution can be written as
\begin{equation}
    \mathbf{w}_{MSE}^{(\infty)} = \mathbf{y}\mathbf{\Phi}^\intercal\left(\mathbf{\Phi}\mathbf{\Phi}^\intercal\right)^+ + \mathbf{w}_{MSE}^{(0)}\left(\mathbb{I}_{F} - \mathbf{U}\mathbf{\Lambda}^+\mathbf{\Lambda}\mathbf{U}^\intercal\right)
\end{equation}
Where the quantity $\mathbf{\Lambda}^+\mathbf{\Lambda}$ is a diagonal matrix with zeros on the diagonal corresponding to zero eigenvalues of $\mathbf{\Phi}\mathbf{\Phi}^\intercal$ and ones on the diagonal corresponding to nonzero eigenvalues of $\mathbf{\Phi}\mathbf{\Phi}^\intercal$. The dynamics can be written in terms of the infinite time solution as 
\begin{equation}
    \mathbf{w}_{MSE}^{(t)} = \mathbf{w}_{MSE}^{(\infty)} + \left(\mathbf{w}_{MSE}^{(0)} - \mathbf{w}_{MSE}^{(\infty)}\right)\mathbf{U}\left(\mathbb{I}_F - \gamma\mathbf{\Lambda}\right)^t\mathbf{U}^\intercal
\end{equation}
Which is Eq. \ref{eqn:mse-dynamics-solution} of the main text.

\subsection{Derivation of high temperature softmax cross entropy dynamics}
In the case of cross entropy, we take the model outputs $\hat{y}$ to be the probability assigned to class $i$ on examples $j$, given by the matrix $\hat{y}_{ij} = \left[\mathbf{P_M}\right]_{ij}$. For our models, we have $\mathbf{P_M}\in\mathbb{R}^{C\times N}$ given by the softmax function applied to the model outputs
\begin{equation}
    \mathbf{P_M} = \frac{e^{\beta\mathbf{w}_{XENT}\mathbf{\Phi}}}{\mathbbm{1}_C^\intercal e^{\beta\mathbf{w}_{XENT}\mathbf{\Phi}}}
\end{equation}
Where $\mathbf{w}_{XENT}\in\mathbb{R}^{C\times F}$ is a weight matrix of learnable parameters, $\beta$ is the inverse temperature, and $\mathbbm{1}_C$ is a $C$ dimensional vector of ones. Exponentials should be interpreted as elementwise operations. We also have the label matrix
\begin{equation}
    \left[\mathbf{P_L}\right]_{ij} = \delta_{i,\argmax{y_j}}
\end{equation}
Where $\delta_{ij}$ is the Kronecker delta. $\mathbf{P_L}$ is thus a matrix of one-hot vectors defining the labels given by the $y_j$s. We also define the label-smoothed label matrix as
\begin{equation}
    \mathbf{\tilde{P}_L}= \alpha\mathbf{P_L} + (1-\alpha)\frac{\mathbbm{1}_{CN}}{C}
\end{equation}
Where $\mathbbm{1}_{CN}$ is a $C\times N$ matrix of ones, and $\alpha$ is a smoothing parameter\cite{szegedy2016rethinking, muller2019does} that obeys $0\leq\alpha\leq1$. Given an $\alpha$, we find the parameters $\mathbf{w}_{XENT}$ by gradient descent on the cross entropy cost defined by 

\begin{equation}
    \mathcal{L}_{XENT}=-\frac{1}{N}\Tr\left[\mathbf{\tilde{P}_L}^\intercal\log\left(\mathbf{P_M}\right)\right]
\end{equation}
Where the log is elementwise. The gradient of $\mathcal{L}_{XENT}$ with respect to the weights $\mathbf{w}_{XENT}$ is given by
\begin{equation}
        \frac{\partial\mathcal{L}}{\partial\mathbf{w}} = \frac{\beta}{N}\left(\mathbf{P_M} - \mathbf{\tilde{P}_L}\right)\mathbf{\Phi}^\intercal
\end{equation}
And the discrete time gradient descent dynamics of $\mathbf{w}_{XENT}$ (abbreviated as $\mathbf{w}$) is given by the nonlinear recursion relation
\begin{equation}
    \label{eqn:nonlinear-recursion}
    \mathbf{w}^{(t+1)} = \mathbf{w}^{(t)} - \frac{\gamma\beta}{N}\left(\mathbf{P_M} - \mathbf{\tilde{P}_L}\right)\mathbf{\Phi}^\intercal
\end{equation}
For learning rate $\gamma$. These dynamics are not solvable in general, but we can gain insight into their behavior by analyzing the high temperature limit. We construct this by taking the first order taylor expansion of $\mathbf{P_M}$ about $\beta=0$, which is
\begin{equation}
    \mathbf{P_M} \approx \frac{\mathbbm{1}_{CN}}{C} + \frac{\beta}{C}\mathbf{M}\mathbf{w}\mathbf{\Phi} +\mathcal{O}\left(\beta^2\right)
\end{equation}
Here $\mathbf{M}$ is defined as
\begin{equation}
    \mathbf{M} \equiv \mathbb{I}_C - \frac{1}{C}\mathbbm{1}_{CC}
\end{equation} 
Where $\mathbbm{1}_{CC}$ is a $C\times C$ matrix of ones, and $\mathbb{I}_C$ is a $C\times C$ identity matrix. Under this approximation, the dynamics simplify to
\begin{equation}
    \mathbf{w}^{(t+1)} = \mathbf{w}^{(t)} - \frac{\gamma\beta^2}{NC}\left[ \mathbf{M}\mathbf{w}^{(t)}\mathbf{\Phi} - \frac{\alpha}{\beta}\left(C\mathbf{P_L} -\mathbbm{1}_{CN}\right)\right]\mathbf{\Phi}^\intercal
\end{equation}
For simplicity, we absorb the factor of $\beta^2/C$ into the learning rate, and take $\beta=\alpha$ under the assumption that $\alpha\ll 1$ to get
\begin{equation}
    \mathbf{w}^{(t+1)} = \mathbf{w}^{(t)} - \frac{\gamma}{N}\left[ \mathbf{M}\mathbf{w}^{(t)}\mathbf{\Phi} - \left(C\mathbf{P_L} -\mathbbm{1}_{CN}\right)\right]\mathbf{\Phi}^\intercal
\end{equation}
Lastly, we make use of two special properties of the matrix $\mathbf{M}$:
\begin{equation}
    \begin{split}
    \mathbf{M}^2 &= \mathbf{M} \\
    \mathbf{M}\left(C\mathbf{P_L} -\mathbbm{1}_{CN}\right) &= \left(C\mathbf{P_L} -\mathbbm{1}_{CN}\right)
    \end{split}
\end{equation}
Both of these follow from the action of $\mathbf{M}$ to subtract the mean across classes of whatever it acts on, and the second follows from the fact that $\left(C\mathbf{P_L} -\mathbbm{1}_{CN}\right)$ is mean zero across classes. Multiplying through the recursion relation with $\mathbf{M}$ and using these properties gives
\begin{equation}
    \mathbf{M}\mathbf{w}^{(t+1)} = \mathbf{M}\mathbf{w}^{(t)} - \frac{\gamma}{N}\mathbf{M}\mathbf{w}^{(t)}\mathbf{\Phi}\mathbf{\Phi}^\intercal + \frac{\gamma}{N}\left(C\mathbf{P_L} -\mathbbm{1}_{CN}\right)\mathbf{\Phi}^\intercal
\end{equation}
These dynamics are identical to the MSE dynamics given in the preceding section with labels $\mathbf{y}=\left(C\mathbf{P_L} -\mathbbm{1}_{CN}\right)$, and can be solved the same way except with a weight matrix given by $\mathbf{M}\mathbf{w}$. This correspondence generates the solutions given in the main text. 

\subsection{Relationship between dynamics of softmax cross entropy and MSE}

As shown in the previous two sections, the training dynamics of a highly label-smoothed softmax cross entropy cost is similar to the dynamics of training with a mean squared error cost. However, differences in the learned weights may occur in the mean across classes. Here we show that in the case of softmax cross entropy, the mean across classes of the weight matrix remaines fixed at its initial value, while in the case of the mean squared error, it decays away to zero. First, we reparameterize the weight matrix via
\begin{equation}
    \begin{split}
    \mathbf{w} &= \mathbf{M}\mathbf{w} + \frac{1}{C}\mathbbm{1}_{CC}\mathbf{w}\\
    &=\tilde{\mathbf{w}} + \mathbf{\mu}\\
    \end{split}
\end{equation}
Where $\tilde{\mathbf{w}}$ has zero mean across classes, and $\mathbf{\mu}$ is the mean of $\mathbf{w}$ across classes. It is clear then that $\mathbf{M}\tilde{\mathbf{w}} = \tilde{\mathbf{w}}$ and $\mathbf{M}\mathbf{\mu}=0$ by the action of $\mathbf{M}$. In this parameterization, the high temperature dynamics of softmax cross entropy training become
\begin{equation}
    \begin{split}
    \tilde{\mathbf{w}}_{XENT}^{(t+1)} &= \tilde{\mathbf{w}}_{XENT}^{(t)} - \frac{\gamma}{N}\tilde{\mathbf{w}}_{XENT}^{(t)}\mathbf{\Phi}\mathbf{\Phi}^\intercal + \frac{\gamma}{N}\left(C\mathbf{P_L} -\mathbbm{1}_{CN}\right)\mathbf{\Phi}^\intercal \\
    \mathbf{\mu}_{XENT}^{(t+1)} &= \mathbf{\mu}_{XENT}^{(t)} \\
    \end{split}
\end{equation}
For MSE, the situation is similar with the exception that the mean $\mathbf{\mu}_{XENT}$ couples to $\mathbf{\Phi}\mathbf{\Phi}^\intercal$. Acting on both sides of the MSE recursion relation with $\mathbf{M}$ gives a recursion relation for $\tilde{\mathbf{w}}_{MSE}$, and a recursion relation for $\mathbf{\mu}_{MSE}$ can be obtained from $\mathbf{\mu} = \mathbf{w} - \tilde{\mathbf{w}}$. The result is 
\begin{equation}
    \begin{split}
    \tilde{\mathbf{w}}_{MSE}^{(t+1)} &= \tilde{\mathbf{w}}_{MSE}^{(t)} - \frac{\gamma}{N}\tilde{\mathbf{w}}_{MSE}^{(t)}\mathbf{\Phi}\mathbf{\Phi}^\intercal + \frac{\gamma}{N}\left(C\mathbf{P_L} -\mathbbm{1}_{CN}\right)\mathbf{\Phi}^\intercal \\
    \mathbf{\mu}_{MSE}^{(t+1)} &= \mathbf{\mu}_{MSE}^{(t)} -\frac{\gamma}{N}\mathbf{\mu}_{MSE}^{(t)}\mathbf{\Phi}\mathbf{\Phi}^\intercal\\
    \end{split}
\end{equation}
Which is identical to the dynamics of high temperature softmax cross entropy with the exception that the mean across classes decays to zero with a rate determined by $\mathbf{\Phi}\mathbf{\Phi}^\intercal$. In general, the argmax model output is independent of $\mathbf{\mu}$, and so in all cases the top-1 accuracy of training with cross entropy and a high-temperature softmax and MSE will be identical.

\subsection{Derivation of expected test cost}
We begin with the solution to the training dynamics in the case of MSE training, which behaves equivalently to high-temperature softmax cross entropy training, and take the special case $\mathbf{\Phi}=\mathbf{X}\sim\mathcal{N}(0, 1)$ and $\mathbf{w}^{(0)}=0$ for $\mathbf{X}\in\mathbb{R}^{N\times D}$ and $\mathbf{w}\in\mathbb{R}^{C\times D}$:
\begin{equation}
    \mathbf{w}^{(t)} = \mathbf{y}\mathbf{X}^\intercal\left(\mathbf{X}\mathbf{X}^\intercal\right)^+  - \mathbf{y}\mathbf{X}^\intercal\left(\mathbf{X}\mathbf{X}^\intercal\right)^+\mathbf{U}\left(\mathbb{I}_D - \gamma\mathbf{\Lambda}\right)^t\mathbf{U}^\intercal
\end{equation}
Taking the labels $\mathbf{y} = \mathbf{w}_T\mathbf{X}_{TRAIN} +\mathbf{\epsilon}$ for a teacher matrix $\mathbf{w}_T\sim\mathcal{N}(0, 1/\sqrt{D})$ and noise $\mathbf{\epsilon}$, we change to the eigenbasis using the SVD $\mathbf{X}/\sqrt{N} = \mathbf{U}\mathbf{\Lambda}^{1/2}\mathbf{V}^\intercal$. This involves defining $\mathbf{q}=\mathbf{w}\mathbf{U}$, $\mathbf{q}_T=\mathbf{w}_T\mathbf{U}$, and $\mathbf{\eta}=\mathbf{\epsilon}\mathbf{V}$. The result is
\begin{equation}
    \mathbf{q}^{(t)} = \left(\mathbf{q}_T+\mathbf{\eta}\mathbf{\Lambda}^{-1/2}\right)\left(\mathbb{I}_{D} - \left(\mathbb{I}_F - \gamma\mathbf{\Lambda}\right)^t\right)
\end{equation}
We can then evaluate the expected generalization performance on noise free test data $x_T\sim\mathcal{N}(0, \mathbb{I}_D)$ via
\begin{equation}
    \left<\mathcal{L}_{test}\right>_{\epsilon, \mathbf{X},x_T, \mathbf{w}_T} = \frac{1}{2}\left<\left(\mathbf{w}x_T - \mathbf{w}_Tx_T\right)^\intercal\left(\mathbf{w}x_T - \mathbf{w}_Tx_T\right)\right>_{\epsilon, \mathbf{X},x_T, \mathbf{w}_T}
\end{equation}
Here the expectation is taken over the distribution of the training data $\mathbf{X}$, the test data $x_T$, the teacher matrix $\mathbf{w}_T$, and the label noise $\mathbf{\epsilon}$. The expectation over $x_T$ is easy, as $\left<x_T^\intercal x_T\right>_{x_T}=\mathbb{I}_D$. Since the MSE is invariant to orthogonal transformations, we can compute the expectations in the rotated basis
\begin{equation}
    \left<\mathcal{L}_{test}\right>_{\epsilon, \mathbf{X}, \mathbf{w}_T} = \frac{1}{2}\left<\left(\mathbf{q} - \mathbf{q}_T\right)^\intercal\left(\mathbf{q} - \mathbf{q}_T\right)\right>_{\eta, \mathbf{\Lambda}, \mathbf{q}_T}
\end{equation}
This can be evaluated for each dimension identically as so far everything is isotropic. Using the solution to the dynamics, and dropping terms first order in $\mathbf{q}_T$ and $\eta$ since $\left<\mathbf{q}_T\right>=0$ and $\left<\eta\right>=0$ gives
\begin{equation}
    \left<\mathcal{L}_{test}\right>_{\epsilon, \mathbf{X}, \mathbf{w}_T} = \frac{1}{2}\left<\left(q_T^2+\frac{\eta^2}{l}\right)(1-\gamma l)^{2t} +\frac{\eta^2}{l}\left(1 - 2(1-\gamma l)^{t}\right)\right>_{\eta, l, q_T}
\end{equation}
Where $q_T$ is a single component of $\mathbf{q}_T$, $\eta$ is a single component of the noise, and $l$ is the corresponding eigenvalue. Using $\left<q_T^2\right>=1$, along with the assumption for the noise
\begin{equation}
   \eta \sim    \begin{cases}
      \mathcal{N}(0, \sigma) & l \geq 1 \\
      0 & l \leq 1\\
    \end{cases}
\end{equation}
We get
\begin{equation}
    \begin{aligned}
    \left<\mathcal{L}_{test}\right>_{\epsilon, \mathbf{X}, \mathbf{w}_T} =& \frac{1}{2}\int_{1}^{\infty}\left[\left(1 + \frac{\sigma}{l}\right)\left(1-\gamma l\right)^{2t} + \frac{\sigma}{l}\left(1 - 2(1-\gamma l)^{t}\right)\right]p(l)dl\\
    &+ \frac{1}{2}\int_{0}^{1}\left[(1-\gamma l)^{2t} \right]p(l)dl
    \end{aligned}
\end{equation}
Here, $p(l)$ is the distribution of eigenvalues of the training data. The expectation over the training data has thus been replaced with an expectation over $l$. Owing to our assumption that the training data is drawn from an uncorrelated gaussian, the eigenvalues of the data covariance matrix $\mathbf{XX^\intercal}/N$ follow the Marchenko-Pasteur (MP) distribution in the limit $N, D \to\infty$ with $D/N=\lambda$.
\begin{equation}
    p_{MP}(x\vert\lambda) = \frac{1}{2\pi}\frac{\sqrt{(\lambda_+ - x)(x - \lambda_-)}}{\lambda x}\mathbf{1}_{x\in\left[\lambda_-, \lambda_+\right]}
\end{equation}
Taking this high-dimensional limit, we get
\begin{equation}
    \begin{aligned}
    \left<\mathcal{L}_{test}\right>_{\epsilon, \mathbf{X}, \mathbf{w}_T} =& \frac{1}{2}\int_{1}^{\infty}\left[\left(1 + \frac{\sigma}{l}\right)\left(1-\gamma l\right)^{2t} + \frac{\sigma}{l}\left(1 - 2(1-\gamma l)^{t}\right)\right]p_{MP}(l\vert\lambda)dl\\
    &+ \frac{1}{2}\int_{0}^{1}\left[(1-\gamma l)^{2t} \right]p_{MP}(l\vert\lambda)dl
    \end{aligned}
\end{equation}
Which is Eq. \ref{eqn:generalization-error} of the main text.

\subsection{Correspondence between label noise and type of noise used in model}
Our assumption that the noise in the training labels couples only to large eigenvalue features is key to obtaining an epochwise double descent effect. Other assumptions, such as noise that couples to all eigenvalues equally do not show epochwise double descent anywhere in the $\lambda, \sigma$ plane. Here we informally explain some of the intuition behind this assumption.

First, we note that noise introduced by randomly shuffling labels (as in the experiments on Deep Networks that show epochwise double descent) does not couple to all features equally in the eigenbasis. To see this, note that we can write totally random labels $\mathbf{y}_{rand}$ in terms of the clean labels $\mathbf{y}_{clean}$ and a noise term $\eta$.
\begin{equation}
    \mathbf{y}_{rand} = \mathbf{y}_{clean} + \eta
\end{equation}
Hence, noise in a dataset with totally random labels can be written as $\eta = \mathbf{y}_{rand} - \mathbf{y}_{clean}$. In a dataset where only a fraction $\epsilon$ of the $N$ training examples have random labels, we introduce the matrix $\mathbf{F}$ where $\mathbf{F}\in\mathbb{R}^{N\times N}$ is a diagonal matrix with $N\epsilon$ ones randomly distributed on the diagonal, and $1-\epsilon$ zeros in the remaining positions on the diagonal. Then we have
\begin{equation}
    \eta = \left(\mathbf{y}_{rand} - \mathbf{y}_{clean}\right)\mathbf{F}
\end{equation}
In the solution to the dynamics, noise enters via the product $\eta\mathbf{X}^\intercal$ for $\mathbf{X}\in\mathbb{R}^{D\times N}$, and so through $\mathbf{F}\mathbf{X}^\intercal$. Using the SVD $\mathbf{X}=\mathbf{U}\mathbf{\Lambda}^{1/2}\mathbf{V}^\intercal$, we can write
\begin{equation}
    \mathbf{F}\mathbf{X}^\intercal = \mathbf{V}\mathbf{F}\mathbf{\Lambda}^{1/2}\mathbf{U}^\intercal - \left[\mathbf{F}, \mathbf{V}\right]\mathbf{\Lambda}^{1/2}\mathbf{U}^\intercal
\end{equation}
The first term on the RHS includes a product $\mathbf{F}\mathbf{\Lambda}^{1/2}$ which acts to drop a subset of the eigenvalues. The second term arises from the commutator $\left[\mathbf{F}, \mathbf{V}\right]$ which has a simple structure:
\begin{equation}
    \left[\mathbf{F}, \mathbf{V}\right] = \mathbf{V}\odot\mathbf{D}
\end{equation}
Where $\mathbf{D} $ is given by $\mathbf{D}_ij = \left(\mathbf{F}_{ii} - \mathbf{F}_{jj}\right)\in\{-1, 0, 1\}$. Thus, the action of fractional label permutation has two properties:
\begin{enumerate}
    \item Some eigenvectors do not couple to the noise (for $D > N$)
\end{enumerate}
\subsection{Compute requirements}
\label{sec:compute}
All of the experiments described here were run on an internal compute cluster. Each experiment was run on a compute unit with 1 GPU, 12 CPU Cores and 16 GB RAM. The results in the main paper stem from ~340 ResNet18 training runs, which consume the bulk of our overall compute usage.

Our experiments were developed using the PyTorch library (pytorch.org), which is open-sourced under the BSD license. Many of our experiments were run on the CIFAR10 dataset, which is publicly available under the MIT license.

\subsection{Experiment hyperparameters}
\label{sec:hparams}
Experiments in Section \ref{sec:width-sweep} were performed using hyperparameters found in \cite{nakkiran2019deep}. This was done to ease comparisons with prior work, but we found those hyperparameters to be suboptimal. In Sections \ref{sec:pca-experiment} and \ref{sec:retrain-experiment} we instead performed a hyperparameter sweep over learning rate, learning rate decay, batch size and momentum. For learning rate decay, we used a schedule of $\gamma(t) = \frac{1}{\beta t + 1}$, evaluated at the end of each epoch. We found the following hyperparameters to work best:
\begin{table}[h!]
    \caption{Experiment hyperparameters}
    \label{tab:hyperparams}
    \centering
    \begin{tabular}{c|c|c}
        \toprule
          & Cross entropy & MSE  \\
        \midrule
         Optimizer & SGD  & SGD \\
         Batch size & 16 & 16 \\
         Learning rate ($\gamma$) & 0.1 & 0.1 \\
         Learning rate decay ($\beta$) & 1 & .01 \\
         Momentum & 0 & 0 \\
    \end{tabular}
\end{table}

\end{document}